\documentclass[conference]{IEEEtran}
\IEEEoverridecommandlockouts
\usepackage{amsmath}
\usepackage{textcomp}
\usepackage{gensymb}
\usepackage{array}
\usepackage{mdwmath}
\usepackage{mdwtab}
\usepackage{eqparbox}
\usepackage{multirow}
\usepackage{stmaryrd}
\usepackage{paralist}
\usepackage{dsfont}
\usepackage{verbatim}
\usepackage{calc}
\usepackage{amssymb}
\usepackage{amstext}
\usepackage{amsthm}
\usepackage{multicol}
\usepackage[small]{caption}
\usepackage{xcolor}
\usepackage{rotating}
\usepackage{subcaption}

\usepackage{algorithm}
\usepackage{algorithmic}
\usepackage[numbers,sort&compress]{natbib}

\usepackage{hyperref}

\usepackage{xcolor}
\usepackage{pgfplots,pgfplotstable}
\usepackage{tikz}
\usepackage{lipsum,adjustbox}
\usetikzlibrary{calc,positioning}
\pgfplotsset{compat=newest}
\newcommand{\ie}[0]{\textit{i.e.},~}
\newcommand{\eg}[0]{\textit{e.g.},~}

\newcommand{\etal}[0]{\textit{et al.}~}

\usepackage{lineno,hyperref}
\modulolinenumbers[5]

\begin{document}

\title{Super-resolution of Omnidirectional Images Using Adversarial Learning}

\author{\IEEEauthorblockN{Cagri Ozcinar*, Aakanksha Rana*, and Aljosa Smolic} \thanks{* are equal contributors.} \thanks{978-1-7281-1817-8/19/\$31.00 ©2019 European Union}
\IEEEauthorblockA{V-SENSE, School of Computer Science and Statistics, Trinity College Dublin, Ireland. }	} 
\maketitle	


\begin{abstract}
An omnidirectional image (ODI) enables viewers to look in every direction from a fixed point through a head-mounted display providing an immersive experience compared to that of a standard image. Designing immersive virtual reality systems with ODIs is challenging as they require high resolution content. In this paper, we study super-resolution for ODIs and propose an improved generative adversarial network based model which is optimized to handle the artifacts obtained in the spherical observational space. Specifically, we propose to use a fast PatchGAN discriminator, as it needs fewer parameters and improves the super-resolution at a fine scale. We also explore the generative models with adversarial learning by introducing a spherical-content specific loss function, called 360-SS. To train and test the performance of our proposed model we prepare a dataset of 4500 ODIs. Our results demonstrate the efficacy of the proposed method and identify new challenges in ODI super-resolution for future investigations.
\end{abstract}


\section{Introduction}\label{intro}

With recent advances in virtual reality (VR), the omnidirectional image (ODI) represents an increasingly important imaging format for immersive technologies. This format provides realistic VR experiences in which viewers have a sense of being present where the content was captured. Existing VR displays, such as a head-mounted displays (HMD), navigate a given ODI input with three degrees of freedom. ODIs are captured with 360\textdegree{} camera systems and stored in 2D planar representations (\eg equirectangular projection~(ERP) or cubemap projection)~\cite{jvetProjections} to be compatible with existing image processing pipelines. For display, the ODI is projected onto a sphere and then rendered through a VR display. When using this emerging technology, viewers receive a more immersive experience compared to watching a standard 2D image. ODIs are used in a variety of applications such as entertainment~\cite{Rana2019}, advertising, communication~\cite{ozcinar19}, health-care, and education.

However, this image representation introduces significant technical challenges because it requires very high resolution to cover the entire 360\textdegree{} viewing space. Existing HMDs can use only part of a given ODI, called the viewport~\cite{myIcip2017}, so even higher resolutions are required to provide an acceptable level of quality of experience (QoE)~\cite{Croci2019} in VR. For instance, according to recent studies~\cite{8329628}, the resolution of the captured ODI should be 21600$\times$10800 (\ie 60 pixels per degree) to provide a sufficiently high-quality ODI for existing HMD devices. At this level, neither current consumer capturing systems nor network bandwidths are capable of processing and transmitting such a large amount of data.

High-resolution content using a single ODI at a lower resolution can be reconstructed with super-resolution (SR) techniques. Early studies such as~\cite{plenopticSR2010,omnisr_tip2011} have shown that high-resolution ODIs with high-quality could be generated using handcrafted image features. A high-resolution ODI was constructed by interpolating the missing information between low-resolution ODI pixels.

In today's era of deep learning, the core SR algorithms for traditional images have shifted from interpolation to learning\cite{RanaTmm} for constructing the high-resolution image from a given low-resolution image using predictive or generative models to find better optimal solutions~\cite{ranammsp,rana-icip17}. Convolution neural networks (CNNs) have shown great successes in the SR field~\cite{Blau2018jr}. In particular, for standard images, high-quality reconstruction performance can be achieved by using recent high-capacity deep learning algorithms such as the generative adversarial networks (GANs)~\cite{Zhang2018-rb,Ledig2017-cw,Wang_Xintao,JMartineau2018mi}. 

In this paper, we address the problem of SR with deep learning for ODIs by introducing an efficient CNN-based solution for reconstructing ODIs with high resolution from given their low-resolution ODIs. In particular, we design a GAN model that inputs a low resolution ODI and outputs a corresponding high-resolution image (2$\times$, 4$\times$, 8$\times$) of higher quality compared to what existing algorithms can reconstruct. Following a generative adversarial network paradigm, our model consists of a generator ($G_{sr}$) and a discriminator ($D_{sr}$), which compete with each other. The generator trying to fool the discriminator by producing high-resolution output ODIs for the given low resolution ODIs. The discriminator tries to distinguish between real and synthetically generated ODI pairs. To enhance the performance of this adversarial learning , we also introduce an ODI-structure-preserving loss function, called 360-SS loss. Our approach handles typical artifacts appearing at the polar regions of the conventional ERP format ODIs, and thus preserving more details of the spherical content in the polar and equatorial regions. 

Our contributions in this paper are two fold. Firstly, we explore the super-resolution of ODIs problem using adversarial learning paradigm for the first time, and propose the use of a PatchGAN~\cite{pix2pix2016} discriminator to constrain the generator for predicting high-quality output effectively. Secondly, we introduce the 360-SS loss function to precisely measure the objective quality of the ODI in spherical observation space. In this loss function, the error of each pixel on the projection plane is multiplied by a weight to account for the influence of the spherical mapping on the loss estimation. The code for the proposed method is provided with this paper\footnote{\url{https://github.com/V-Sense/360SR}}. 
We hope that our work can inspire further research within the context of deep learning-based SR to reconstruct high-resolution ODIs.

The rest of this paper is organized as follows. Section~\ref{related-work} presents the related work on super-resolution, focusing on early work for ODIs and recent deep learning-based approaches. The proposed model and experimental results are presented in Sections~\ref{model} and~\ref{results} followed by our conclusions in Section~\ref{conclusion}.

\begin{figure*}[h]
    \centering
    \includegraphics[width=\linewidth]{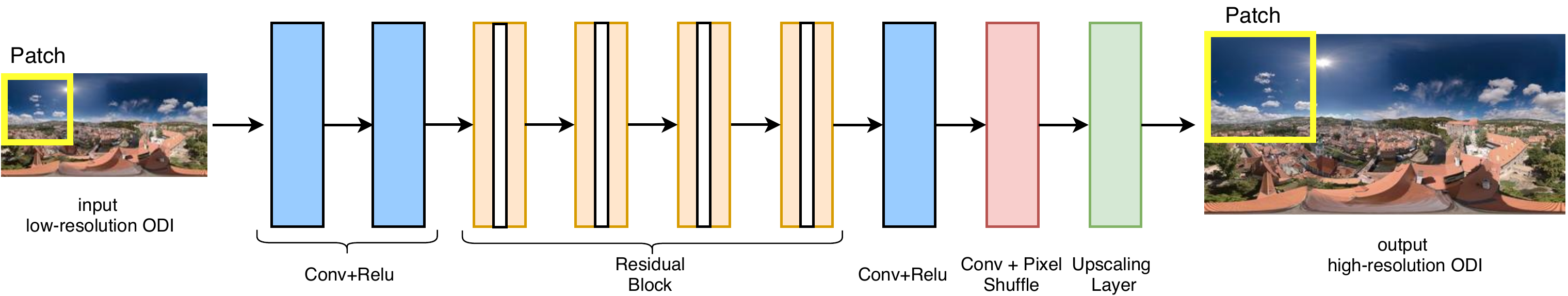}
  \caption{Architecture of Generator Network with corresponding block labels.}
  \label{fig:model}
\end{figure*}

\section{Related Work}\label{related-work}

In previous two decades, many super-resolution methods have been proposed, and we briefly describe them in this section. For more details and examples from this field, comprehensive literature reviews are available in~\cite{nasrollahi2014super, Wang2019-qi}.

Early work, such as~\cite{plenopticSR2010,omnisr_tip2011}, proposed innovative super-resolution techniques that utilized handcrafted methods or simplistic learning based techniques~\cite{2014feature} for ODIs. In~\cite{plenopticSR2010}, for instance, the authors first performed a registration between successive frames of omnidirectional video using the plenoptic geometry of the 360-degree scene. All the visual information was then used to generate a high-resolution ODI. Later, Arican~\etal~\cite{omnisr_tip2011} showed how multiple ODIs with arbitrary rotations could contribute to the super-resolution problem by leveraging the Spherical Fourier Transform (SFT). In their work, the joint registration and super-resolution problem was solved based on the total least squares norm minimization in the SFT domain.

Recent image super-resolution efforts benefited from advances in deep learning~\cite{ranacolor} and end-to-end architectures. For example, Zhang~\etal~\cite{Zhang2018-rb} introduced a dimensional stretching strategy for a single deep-learning network to handle multiple quality degradations (\ie blur kernel and noise level) for super-resolution. Furthermore, Ledig~\etal~\cite{Ledig2017-cw} highlighted the limitations of PSNR-based super-resolution solutions and introduced a generative network for producing super-resolutions, called SRGAN. In their work, a perceptual loss was introduced to optimize the super-resolution model in a future space instead of pixel space. Later, to recover more realistic textural details, Wang~\etal~\cite{Wang_Xintao} improved the discriminator of the SRGAN using a realistic average GAN~\cite{JMartineau2018mi}. Additionally, they enhanced the perceptual loss by incorporating features before activation.

However, super-resolution for ODI has not been investigated sufficiently. To the best of our knowledge, no research work on the super-resolution of ODIs using recent advances in deep learning exist.





\section{Proposed Model}
\label{model}

Our objective is to generate a high-resolution ODI, $I_{360-sr}$, from a low-resolution input ODI, $I_{360-lr}$. To this end, we first prepared the ODI pairs, the output $I_{360-sr}$ and input $I_{360-lr}$, to be used for training. We obtained the low resolution version following the standard methodology mentioned in~\cite{Ledig2017-cw}, where a Gaussian filter followed by a down-sampling operator is applied to obtain the low-resolution ODIs. With these training image-pairs, we then propose to learn the generator $\mathbf{G}$ to generate artifact-free and a better quality high-resolution output images by optimizing the whole network with an adequate loss functions. In this paper, we employ the most commonly used ERP ODI representation. 

In the following, we explain the proposed network architecture including the generator ($\mathbf{G}$) and discriminator ($\mathbf{D}$) and the proposed loss function (360-SS) to preserve the spherical content specific details. 

\begin{figure}[tb]
    \centering
    \includegraphics[width=\columnwidth]{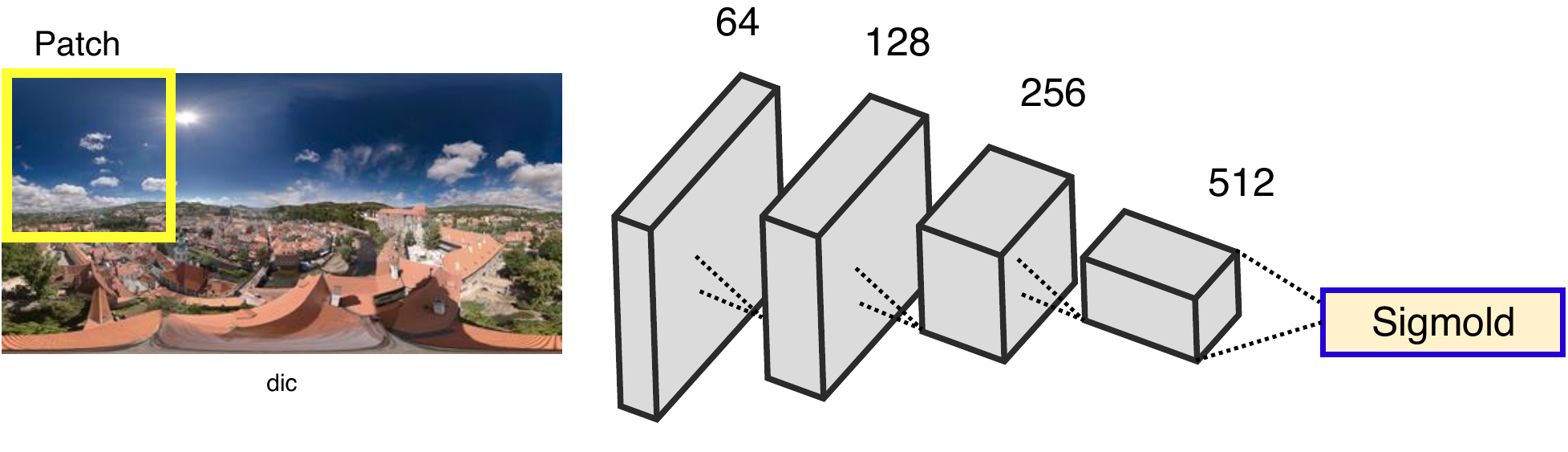}
  \caption{Architecture of the used Discriminator Network.}
  \label{fig:disc}
  \vspace{-4pt}
\end{figure}

\subsection{Generator-Discriminator Architecture}

\textit{Motivation:} The fundamental idea behind a GAN framework is that both the generator and discriminator try to compete with each other. On the one hand, the generator tries to fool discriminator by producing a high resolution, real looking ODI by taking a low-resolution ODI input. While on the other hand, the discriminator tries to discriminate between real and synthetically generated (by the generator) high-resolution ODI image pairs. In this process by spanning the sub-space of the natural images, the generator learns to generate images that are much closer to their corresponding ground truth. This, on the contrary, is quite difficult to achieve with a simple CNN network using a Euclidean loss~\cite{pix2pix2016}. In this paper, inspired from several state-of-art approaches ~\cite{Ledig2017-cw,Wang_Xintao}, we explore the adversarial learning framework for our task at hand. 

\textit{GAN:} In order to design a model for super-resolving ODIs, we adopt the generator architecture from~\cite{Ledig2017-cw}. The layout of the generator architecture is shown in Figure~\ref{fig:model}. The network consists of a convolutional layer followed by five residual blocks which are capable of capturing the fine details with high-frequency, bypassing the need of a deeper convolutional layer network~\cite{Ledig2017-cw}. The network consists explicitly of two convolution layers and a final sub-pixel convolution layer which is designed to super-resolute the input image by the given factor.

Next, we propose to use the PatchGAN discriminator to discriminate between the real and generated ODI. The layout of the discriminator architecture is shown in Figure~\ref{fig:disc}. The PatchGAN discriminator is applied on a $70\times70$ (overlapping) patch-size of its input, and the averaged scores are then used to finalize the decision whether the image is real or fake. In this process, the network learns to focus at finer patch level details. An added advantage of a PatchGAN is the smaller number of the parameters as compared to the discriminator adopted in~\cite{Ledig2017-cw}, which makes the computation faster.

\textit{Objective Function:} 
In this work, our objective is to construct high-resolution ODI with high-quality from a low resolution ODI using super-resolution approach. For this aim, we formulate our objective as follows:

\begin{equation}
\begin{split}
\mathbf{G}^{\ast} = \arg \min_{G} \max_{D} [\mathcal{L}_{adv} (\mathbf{G}, \mathbf{D})] + \\ \beta \mathcal{L}_{feat} (\mathbf{G})
+ \gamma \mathcal{L}_{360-SS} (\mathbf{G})
\end{split}	
\end{equation}

where we aim to optimize the objective over three loss functions which are used in the adversarial learning of the generator $\mathbf{G}$. In the next section we describe the 3 loss terms: adversarial loss $\mathcal{L}_{adv}$, feature (content) loss $\mathcal{L}_{feat}$ and spherical loss $\mathcal{L}_{360-SS}$ . Initially, we detail the proposed loss function (360-SS) to preserve the spherical content specific details. Then, we discuss the feature and adversarial loss terms.

\subsection{Loss Functions}
\paragraph{360-SS Loss}
To account for the distortion of the spherical surface, we apply the weighted-to-spherically-uniform structural similarity~(WS-SSIM)~\cite{ws_ssim} quality score. This measurement considers mapping of spherical content onto the planar surface of a given ODI by adding the appropriate weight to the SSIM value~\cite{SSIM}. As a traditional $l_2$ -based loss function is inferior to SSIM in many imaging problems, and the traditional SSIM~\cite{SSIM} is not suitable for evaluating ODIs, we use a novel WS-SSIM-based loss function, called $l_{360-SS}$. In this loss function, the spherical surface is considered using a non-linear weighting in the SSIM calculation. SSIM relies on computation of luminance, contrast and structural similarities between distorted and original images and is more consistent with subjective quality evaluation than the traditional $l_2$ based loss function~\cite{Ledig2017-cw}. Similar as~\cite{Jvet2016tw}, the non-linear weights are calculated using the stretching ratio of the area that is projected from the planar surface to the spherical surface. Thus, the loss of 360-SS can be formulated as follows:

\begin{equation}
\mathcal{L}_{360-SS} = \frac{1}{K}\sum_{i=1}^{K} d_{360-SS}^{i},
\end{equation}
where $K$ represents the number of samples, and
\begin{equation}
\begin{aligned}
d_{360-SS} = 
\frac{\sum_{x=1}^{W/r} \sum_{y=1}^{H/r}\Big( SSIM\big( I^{x,y}_{360-sr} , \widehat{I}_{360-sr}^{x,y}\big) q_r^{x,y} \Big)}{\displaystyle \sum_{x=1}^{W/r}\sum_{y=1}^{H/r}q_r^{x,y}},
\end{aligned}
\end{equation}
where $W \times H$ is the resolution of the reconstructed version of the ERP ODI. Note that $x$ and $y$ denote the pixel coordinates of the ERP image, $\widehat{I}_{360-sr}$ and $I_{360-sr}$ stand for the generated high-resolution and original high-resolution versions of the ODI, and $q_r^{x,y}$ represents the weighting intensity in ($x$, $y$) of the weight distribution of the ERP which can be calculated according to \cite{Jvet2016tw} with:
\begin{equation}
\label{weightEq}
q_r^{x,y} = cos \frac{(y+0.5-(H/2r)\pi}{(H/r)}.
\end{equation}

\paragraph{Feature Loss}
To penalize the $\mathbf{G}$ for the distortions in the content of ODIs, we incorporate the VGG feature-based loss~\cite{Simonyan14c} function. This loss is computed as the distance between the feature maps $ \mathcal{F} $ obtained by passing the ground truth ODI ($I_{360-sr}$) and the ODI generated by super-resolution ($\widehat{I}_{360-sr}$) through a pre-trained VGG-19~\cite{Simonyan14c} network. The feature loss $\mathcal{L}_{feat}$ is given as

\begin{equation}
    \mathcal{L}_{feat} = \frac{1}{K}\sum_{i=1}^{K}(\mathcal{F}^{i}(I_{360-sr}) - \mathcal{F}^{i}(\widehat{I}_{360-sr})),
\end{equation}

\paragraph{Adversarial Loss}
We apply adversarial loss~\cite{goodfellow2014generative} to aide the generator to effectively generate high-resolution ODIs from the manifold of natural image sets while fooling the discriminator $\mathbf{D}$ network. The adversarial loss term is computed over all the training samples and is given as:
\begin{equation}
    \mathcal{L}_{adv} = \sum_{i=1}^{K} (-log \mathbf{D} (\mathbf{G} (I_{360-lr}^{i}))).
\end{equation}
The adversarial loss is taken as the negative logarithm of the probability of the super-resoluted image, similar to ~\cite{Ledig2017-cw}.

\section{Results \& Discussions}\label{results}
Next, we describe the used dataset, metrics, implementation details, and results obtained by our proposed model.

\begin{figure*}[h]
\includegraphics[width=\linewidth]{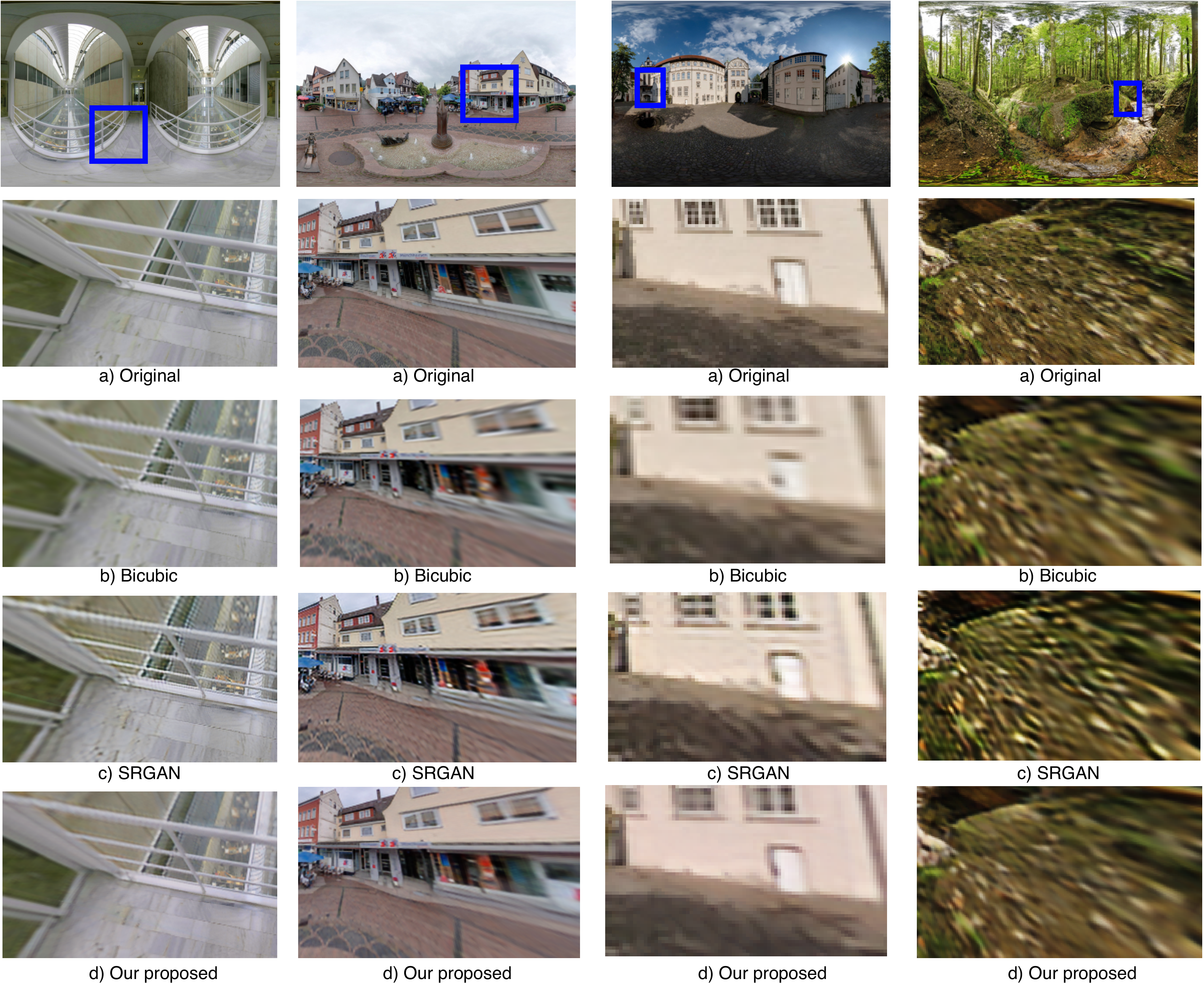}
	\caption{Qualitative reconstruction results obtained with different benchmark methods and corresponding high-resolution original ODI. We applied rectilinear projection to visualize the most significant differences.}
	\label{fig:quality}
\end{figure*}

\subsection{Dataset and Metrics}

\textbf{Dataset}: We randomly selected a total of 4500 ODIs from the SUN 360 Panorama Database~\cite{xiao2012recognizing}. This database is the only publicly available ODI dataset that has many high-resolution panorama images with a wide variety of content, including indoor, outdoor, structures, human faces. Each selected ODI is in ERP format, and the resolution size is $1440\times960$. More specifically, we use the ERP format for training and evaluations. 3500 ODIs were randomly selected from the total set for training purposes, and 500 ODIs were used for validation and test sets. 

\textbf{Metrics}: We evaluated the performance of our proposed model with four metrics: SSIM~\cite{SSIM}, PSNR, WS-SSIM~\cite{ws_ssim}, and WS-PSNR~\cite{WS_PSNR}. SSIM and PNSR are the well-known full-reference planar metrics that have been used in the literature to measure the reconstructed image quality using the original image. To account for the visual distortion with stretching factors on planar representation, we used well-known ODI metrics, S-SSIM and WS-PSNR, to evaluate the performance of our proposed method. Both ODI quality metrics are the extended versions of their traditional metrics by considering the nonlinear relationship between the projection plane and the sphere surface. Subjective experiments in~\cite{Luz2017-ri,Croci2019,Zhang2018-dj,Sun2018-au} show that ODI metrics have high correlation with human votes compared to the standard planar metrics applied to the ERP formats.


\begin{table}[h]
	\centering{
	\caption{Quantitative Results for Super-Resolution by a factor of 2 on 500 ODIs.}
	\label{tab:quantresults1}
	\begin{adjustbox}{width=\linewidth}
		\begin{tabular}{c|cccc}
			\hline
			\multirow{2}{*}{\textbf{Method}} & \multicolumn{4}{c}{\textbf{$r$ = 2}} \\	
 			\cline{2-5}
             & SSIM & PSNR  & WS-SSIM & WS-PSNR \\ 
			\hline
			NN	 &  0.92 $\pm$ 0.06 &  29.38 $\pm$ 0.04 & 0.86$\pm$ .03  & 34.34 $\pm$ .05\\
			Bicubic &  0.93 $\pm$ 0.05 & 30.64 $\pm$ 0.06 & 0.88 $\pm$ .04 & 35.54 $\pm$ .07\\

			SRGAN~\cite{Ledig2017-cw} &  0.94 $\pm$ 0.05 &  32.56 $\pm$ 0.06 & 0.90 $\pm$ .06& 36.35 $\pm$ .06\\
			Ours  &  0.95 $\pm$ 0.04 &  33.20 $\pm$ 0.04 & 0.92 $\pm$ .04 & 37.68 $\pm$ .05 \\	
			\hline
			\textbf{Ours+ 360-SS loss} &  0.95 $\pm$ 0.03 &  33.56 $\pm$ 0.04 & 0.93 $\pm$ .06 & 37.96 $\pm$ .03\\	
			\hline
		\end{tabular}
		\end{adjustbox}
	}    

\end{table}

\begin{table}[h]
	\caption{Quantitative Results for Super-Resolution by a factor of 4 on 500 ODIs.} 
    \label{tab:quantresults_2}
	\centering{
	\begin{adjustbox}{width=\linewidth}
		\begin{tabular}{c|cccc}
			\hline
			\multirow{2}{*}{\textbf{Method}} & \multicolumn{4}{c}{\textbf{$r$ = 4}} \\	
 			\cline{2-5}
             & SSIM & PSNR  & WS-SSIM & WS-PSNR \\
			\hline
			NN	 &  0.83 $\pm$ 0.07 &  25.77 $\pm$ 0.05  & 0.71$\pm$ .03  & 32.44 $\pm$ .05\\
			Bicubic &  0.85 $\pm$ 0.07 & 26.71 $\pm$ 0.03  & 0.74 $\pm$ .04 & 32.76 $\pm$ .04\\

			SRGAN~\cite{Ledig2017-cw} &  0.86 $\pm$ 0.02 &  27.11 $\pm$ 0.09  & 0.75 $\pm$ .06& 34.76 $\pm$ .03  \\
			Ours  &  0.87 $\pm$ 0.05 &  27.19 $\pm$ 0.08 & 0.76 $\pm$ .05 & 35.89 $\pm$ .05\\	
			\hline
			\textbf{Ours+ 360-SS loss} &  0.87 $\pm$ 0.04 &  27.70 $\pm$ 0.03 & 0.77 $\pm$ .08  & 36.98 $\pm$ .06\\	
			\hline
		\end{tabular}
	\end{adjustbox}}
	\label{tab:quantresults2}
\end{table}

\begin{table}[h]
	\caption{Quantitative Results for Super-Resolution by a factor of 8 on 500 ODIs.}
	\centering{
	\begin{adjustbox}{width=\linewidth}
		\begin{tabular}{c|cccc}
			\hline
			\multirow{2}{*}{\textbf{Method}} & \multicolumn{4}{c}{\textbf{$r$ = 8}} \\	
 			\cline{2-5}
             & SSIM & PSNR  & WS-SSIM & WS-PSNR \\
			\hline
			NN	 &  0.83 $\pm$ 0.07 &  23.47 $\pm$ 0.05 & 0.64 $\pm$ .06  & 31.12 $\pm$ .04 \\
			Bicubic &  0.85 $\pm$ 0.07 & 24.26 $\pm$ 0.03 & 0.66 $\pm$ .07 &  31.83 $\pm$ .06\\

			SRGAN~\cite{Ledig2017-cw} &  0.86 $\pm$ 0.02 &  25.10 $\pm$ 0.09 & 0.70 $\pm$ .06 & 33.00 $\pm$ .07\\
			Ours  &  0.87 $\pm$ 0.05 &  26.24 $\pm$ 0.08 & 0.73 $\pm$ .07 & 34.68 $\pm$ .04\\	
			\hline
			\textbf{Ours + 360-SS loss} &  0.87 $\pm$ 0.04 &  26.56 $\pm$ 0.03 & 0.75 $\pm$ .06 &  35.54 $\pm$ .06 \\	
			\hline
		\end{tabular}
		\end{adjustbox}}
	\label{tab:quantresults3}
\end{table}

\subsection{Training and Implementation Details} 
Training was performed on a set of 3500 images of resolution size $ 1440\times960 $ each, where ODIs were down-scaled by a required factor $r\in$ $\{2, 4, 8\}$. While training, random crops of size $512\times512$ were applied to the images. Additional data augmentation techniques such as rotation and flipping were applied to scale up the training dataset. 

Our proposed model was implemented using the Pytorch deep learning library~\cite{paszke2017automatic}. To train each model, we set the batch size to $16$. The weights of the all the layers were initialized randomly and the network was trained from the scratch. Further, to optimize the network we used the ADAM solver~\cite{KingmaB14adam2} with learning rates of $10^{-4}$. Both control parameters $\beta$ and $\gamma$ are set equal to 10. All our models were trained in an end-to-end fashion for $100$ epochs. 
Training was done using a $12$ GB NVIDIA Titan-X GPU on an Intel Xeon E7 core i7 machine which took approximately $2$ hours. Inference time is $0.030$ milliseconds for each ODI. Please not that the inference time is same as that of SRGAN due to the similar generator architecture.

\subsection{Quantitative and Qualitative Analysis}

Here, we compare the performance between our proposed method and other state-of-the-art methods using quantitative and qualitative analysis. For comparison, we used well-known interpolation techniques: nearest-neighbor (NN) and bicubic, and the state-of-the-art SR algorithm SR-GAN, which is also the base of our proposed method. Quantitative results are summarized in Tables~\ref{tab:quantresults1}-\ref{tab:quantresults3} and visual examples are provided in Figure~\ref{fig:quality} for qualitative analysis. 

In Tables~\ref{tab:quantresults1}-\ref{tab:quantresults3}, we show mean and variation values for each method over 500 test ODIs. Looking at the tables, with respect to PSNR and SSIM, we see that the scores obtained using our method have a marginal gain with respect to the other methods. However, for ODI metrics, we obtain higher gains. This accounts to the idea of preserving finer details around the equatorial regions as compared to regions near the poles. Additionally, we observe that our model performs better with higher down-sampling factors (\eg 8$\times$). This further validates our idea of using a Patch-GAN based discriminator instead of the more-layered (deeper) discriminator of SRGAN~\cite{Ledig2017-cw}. This analysis can also benefit other high-resolution imagery studies~\cite{Tarabalka14}, such as HDR imaging~\cite{rana-vcip16,rana-icme17}. Furthermore, our proposed loss function facilitates learning of spherically distorted content thereby, resulting in higher gains. 

To provide qualitative visual comparison between methods, Figure~\ref{fig:quality} shows some examples of constructed high-resolution ODIs for \textit{Bicubic}, \textit{SRGAN}, and our \textit{proposed method}. We magnify details to display the most significant differences using rectilinear projection. As can be seen in the results, the proposed method can construct high-resolution ODI with higher-quality, compared to the benchmark methods.

As shown in the experimental results, the proposed approach achieves the highest performance on ERP format. However, other formats (\eg cubemap, equi-angular cube map, pyramid formats, etc.~\cite{jvetProjections}) could also be employed by modifying the Eq.~\ref{weightEq}. 



\section{Conclusion}
\label{conclusion}

In this paper, we studied the super-resolution problem for omnidirectional images (ODIs) and proposed a new generative adversarial network model optimized to handle the artifacts typically obtained in the spherical observation space of ODIs. The proposed method utilized a PatchGAN discriminator to constrain the generator for predicting high-quality ODI output effectively. A novel loss function called 360-SS was also introduced to precisely measure the objective quality of the ODIs in spherical observation space. The proposed method was compared with state-of-the-art super-resolution techniques developed for standard images, and its performance was verified with quantitative and qualitative results. As future work, we plan to improve and apply the proposed method in ODI streaming application scenarios.


\section*{Acknowledgements}
{This publication has emanated from research conducted with the financial support of Science Foundation Ireland (SFI) under the Grant Number 15/RP/2776. We gratefully acknowledge the support of NVIDIA Corporation with the donated GPU used for this research.}
\small
\bibliographystyle{IEEEtran}
\bibliography{ref.bib}

\end{document}